\pdfoutput=1

\documentclass[11pt]{article}

\usepackage{ACL2023}

\usepackage{times}
\usepackage{latexsym}
\usepackage{amsfonts}
\usepackage{amsmath}
\usepackage{multirow}
\usepackage{booktabs}
\usepackage{graphicx}
\usepackage{subfigure}
\usepackage{tabularx}

\usepackage{graphicx}  
\usepackage{caption}   
\usepackage{subcaption}  

\usepackage[T1]{fontenc}

\usepackage[utf8]{inputenc}

\usepackage{microtype}

\usepackage{inconsolata}

\title{Do Large Language Models Understand Logic or Just Mimick Context?}

\author{Junbing Yan$^{1}$, Chengyu Wang$^{2}$, Jun Huang$^{2}$, Wei Zhang$^{1}$\thanks{\ \ \ Correspondence to Wei Zhang.}\\
  $^{1}$ Alibaba Group, Hangzhou, China \\
  $^{2}$ East China Normal University, Shanghai, China \\
  \texttt{\{junbingyan531,zhangwei.thu2011\}@gmail.com} \\
  \texttt{\{chengyu.wcy,huangjun.hj\}@alibaba-inc.com,}}

\begin{document}
\maketitle
\begin{abstract}
Over the past few years, the abilities of large language models (LLMs) have received extensive attention, which have performed exceptionally well in complicated scenarios such as logical reasoning and symbolic inference. A significant factor contributing to this progress is the benefit of in-context learning and few-shot prompting. However, the reasons behind the success of such models using contextual reasoning have not been fully explored. Do LLMs have understand logical rules to draw inferences, or do they ``guess'' the answers by learning a type of probabilistic mapping through context? This paper investigates the reasoning capabilities of LLMs on two logical reasoning datasets by using counterfactual methods to replace context text and modify logical concepts. Based on our analysis, it is found that LLMs do not truly understand logical rules; rather, in-context learning has simply enhanced the likelihood of these models arriving at the correct answers. If one alters certain words in the context text or changes the concepts of logical terms, the outputs of LLMs can be significantly disrupted, leading to counter-intuitive responses.
This work provides critical insights into the limitations of LLMs, underscoring the need for more robust mechanisms to ensure reliable logical reasoning in LLMs.
\end{abstract}

\section{Introduction}
Logical reasoning is a core component of human cognition that is essential for comprehending, interacting with, and influencing our environment. In contrast to artificial intelligence systems that typically depend on vast datasets and substantial training to build skills, humans excel at employing logical reasoning to deduce, troubleshoot, and assimilate new knowledge from limited data or abstract principles. 
Moreover, humans demonstrate an exceptional capacity to derive novel insights from a minimal number of instances or from theoretical frameworks, a capability that stands in sharp contrast to the extensive, supervised datasets necessitated by deep learning algorithms.
Over the past two years, advancements in large language models (LLMs) have led to extraordinary achievements~\cite{gpt3, instructGPT, opp_of_llm, tf_as_eng}. These models have not only excelled in open-ended tasks such as generating creative dialogues, but have also performed exceptionally well in complex problems that necessitate logical reasoning, common sense, and mathematical skills~\cite{instructGPT, zeroshotlearner, self-instruct}, thanks in part to innovations such as in-context learning~\cite{gpt3, rethink_in_context, nlp_ins, meta_icl, cross_task_ins} and Chain-of-Thought (COT) prompting~\cite{cot}.

In the literature, COT~\cite{cot} is designed to improve the performance in mathematical problem solving by using intermediate steps as prompts, thereby incrementally guiding LLMs through the necessary reasoning process. LogicalCOT~\cite{logicalcot} extends this strategy of intermediate prompting to logical reasoning tasks. While these prompting-based methods have enhanced the performance of LLMs on tasks that require logical reasoning, there is still a gap in our understanding of whether these models have genuinely grasped the underlying logical rules, or whether they simply become more effective at converging to the correct answers.

Therefore, the question remains:~\emph{do the observed proficiencies of LLMs stem from true understanding, or do they merely remember the results based on large-scale parameters, extensive pre-training on large corpora, and a plethora of contextual examples that allow for a broader retention of knowledge?} 
To delve into the topic, we establish a comprehensive evaluation framework based on in-context learning.
We first define the texts, the logical reasoning chain, and reasoning keywords in in-context examples.
We test whether larger models exhibit different behaviors on texts that have undergone modifications or deletions of these components.
Furthermore, we add concepts related to logical definitions and test whether the models understand the relationships between these logical terms by replacing the logical concepts.

Through extensive analysis, the main important findings are summarized as follows:

\begin{itemize}
\item \textbf{The Chain of Thought (COT) in-context examples markedly improve the performance of large-scale models on logical reasoning tasks}. Across a range of models with 7 to 200 billion parameters, these examples significantly enhance the clarity, normativity, and accuracy of the generated responses.

\item \textbf{Large models demonstrate resilience to distracting elements within in-context examples, such as extraneous text, reasoning chains, and patterns.} When various segments of the in-context example content are replaced with text from within or outside the domain, large models (70B and 200B parameters) maintain their output accuracy. In contrast, smaller models (7B and 13B parameters) suffer notable declines in performance when standard in-context examples are not used.

\item \textbf{Large models do not genuinely comprehend logical principles; rather, they rely on probabilistic associations between input examples and outputs.} Efforts to alter the definitions of logical symbols and direct the models to revise their outputs accordingly were met with a minimal rate of successful adaptation across all model sizes. Attempts to enhance the rate of successful adjustments using either prompt or in-context guidance yielded limited improvement.
\end{itemize}

\section{Related Work}

\begin{figure*}
\centering
\includegraphics[width=\textwidth]{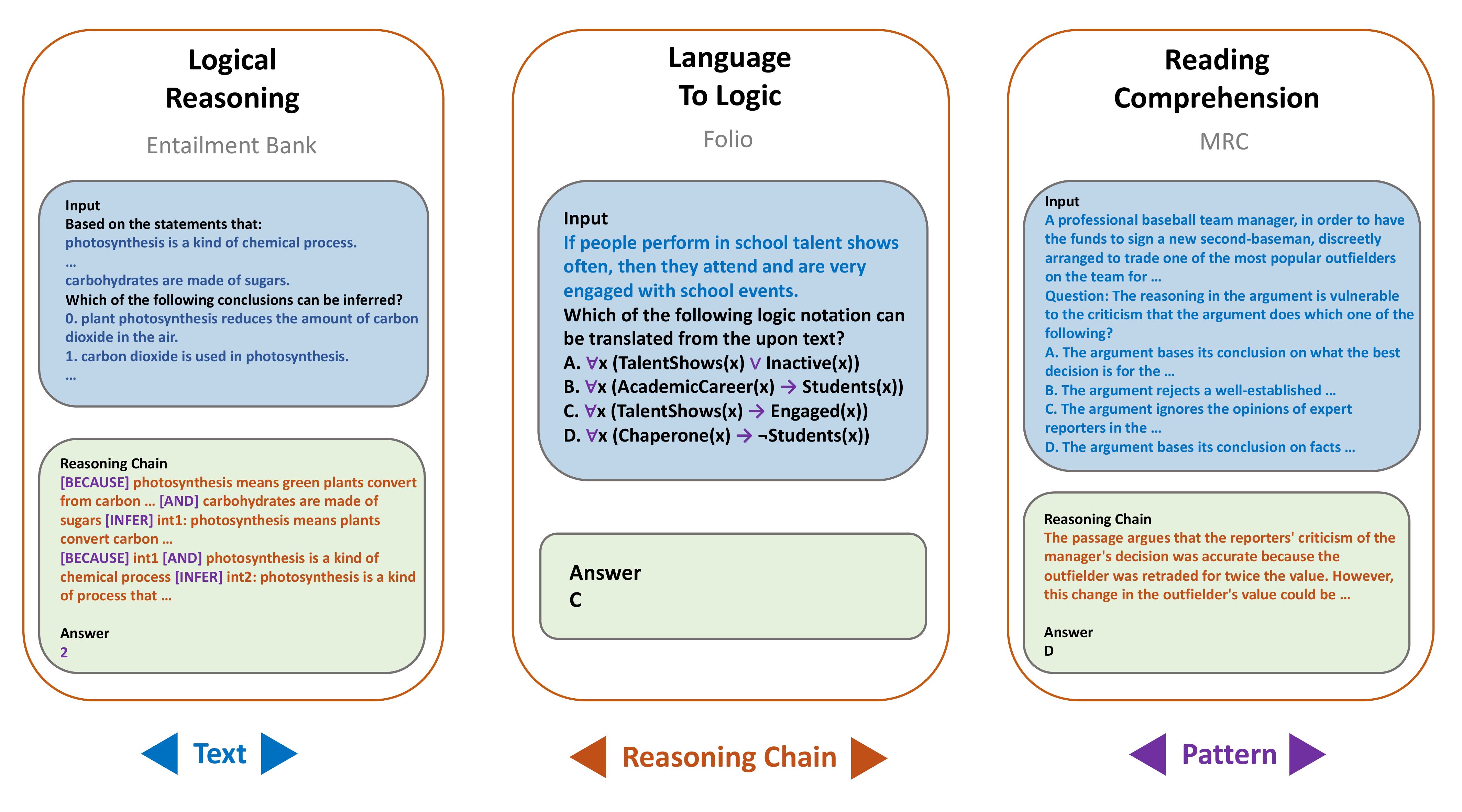}
\caption{Tasks and datasets used in our experiment: \textbf{Text:} in blue color; \textbf{Reasoning Chain:} in orange color; \textbf{Pattern:} in purple color.}
\end{figure*}

\subsection{Large Language Models}
Prior to the emergence of the Large Language Model (LLM) trend, Pre-trained Language Models (PLMs) were already in the spotlight for their proficiency in acquiring contextual representations~\cite{DBLP:journals/corr/abs-2003-08271,DBLP:journals/corr/abs-2111-01243}.
With the escalating size of PLM parameters, there has been a notable enhancement in their performance across a range of NLP tasks, with decoder-only models showing particularly impressive gains. Among these, the 175B-parameter ChatGPT stands out, exhibiting the capacity to craft responses that closely mimic human conversation, leveraging GPT-3's foundational architecture~\cite{DBLP:conf/nips/BrownMRSKDNSSAA20}. Subsequent to the introduction of ChatGPT, the designation "Large Language Model (LLM)" has become commonplace when describing PLMs of considerable scale and exceptional generative capabilities.
Following ChatGPT's launch, the field has seen the advent of numerous LLMs. A selection of prominent open-source LLMs comprises LLaMA~\cite{DBLP:journals/corr/abs-2302-13971}, 
LLaMA 2~\cite{DBLP:journals/corr/abs-2307-09288}, BLOOM~\cite{DBLP:journals/corr/abs-2211-05100}, BLOOMZ~\cite{DBLP:conf/acl/MuennighoffWSRB23}, Galactica~\cite{DBLP:journals/corr/abs-2211-09085},
GLM~\cite{DBLP:conf/iclr/ZengLDWL0YXZXTM23}, Pythia~\cite{DBLP:conf/icml/BidermanSABOHKP23}, among others.
In terms of training methodology, the tripartite framework of "pre-training, supervised fine-tuning (SFT), and Reinforcement Learning from Human Feedback (RLHF)" as proposed by~\cite{DBLP:conf/nips/Ouyang0JAWMZASR22} has gained wide recognition and adoption within the community.

\subsection{Counterfactual Prompt}
A number of recent works have investigated generating counterfactual text in specific language domains (e.g., court view~\cite{wu2020biased}, dialogue generation~\cite{zhu2020counterfactual}, Natural Language Inference~\cite{kaushik2019learning, semantically_robust_optimization}, named entity recognition~\cite{zeng2020counterfactual}). Counterfactual explanations offer a pathway to gain deeper insight into the workings of models. This approach may provide more advantageous interpretations for state-of-the-art Large Language Models (LLMs).

\subsection{Logical Reasoning}
Logical reasoning constitutes a fundamental facet of human cognition and is an essential feature for artificial intelligence systems. To endow AI with this capability, researchers have investigated a multitude of strategies, such as rule-based and symbolic systems \cite{maccartney-manning-2007-natural}, the refinement of expansive language models \cite{wang-etal-2018-glue}, and the integration of neural and symbolic methodologies \cite{li-srikumar-2019-augmenting}. 
Since the introduction of Large Language Models (LLMs) and the development of chain-of-thought prompting~\cite{cot}, there has been a marked enhancement in the logical reasoning capabilities of these models, as evidenced by improved performance metrics across a range of logic tasks. To our knowledge, we are the first to employ counterfactual methods to examine the extent to which these expansive models comprehend logical rules and definitions.

\section{Method}

This study aims to investigate which parts of the in-context examples make a major contribution to the reasoning process of Language Models and whether LLMs understand the reasoning process demonstrated within the examples. To achieve this, we have systematically divided the text within examples into three components: text, reasoning chain, and pattern. Additionally, we have included definitions of logical symbols as supplementary text.
\noindent\textbf{Text:} A sequence of tokens that describe the question to be answered (e.g.,) and the text that contains the given information.
\noindent\textbf{Reasoning Chain:} The thought process regarding the answer to the question, which includes the reasoning pathway pertinent to the current question. 
\noindent\textbf{Pattern:} Key symbols, answers, and other special texts within the in-context examples. 
\noindent\textbf{Definition:} Natural language text providing definitions of logical symbols.

The operations on the aforementioned parts mainly involve two actions: replacement and modification.

\begin{table*}
  \centering
  \begin{small}
  \begin{tabularx}{\textwidth}{l|XX}
    \toprule
    \textbf{} &  \textbf{Origin} & \textbf{After Operation} \\
    \midrule
    \textbf{Text} & Based on the statements that: {\textcolor{blue}{[A set of conditions]}} Which of the following conclusions can be inferred? {\textcolor{blue}{[A set of conditions]}} & Based on the statements that: {\textcolor{blue}{[A set of conditions from other samples]}} / {\textcolor{red}{[A paragragh from Wikipedia]}} Which of the following conclusions can be inferred? {\textcolor{blue}{[A set of conditions from other samples]}} / {\textcolor{red}{[A set of sentences from Wikipedia]}}\\
    \midrule
    \textbf{Chain} & [BECAUSE] {\textcolor{blue}{[statement$_1$]}} [AND] {\textcolor{blue}{[statement$_2$]}} [INFER] {\textcolor{blue}{[Inference$_1$]}} & [BECAUSE] {\textcolor{blue}{[Statement$_1$ from other samples]}} / {\textcolor{red}{[A sentence from Wikipedia]}} [AND] {\textcolor{blue}{[statement$_2$ from other samples]}} / {\textcolor{red}{[A sentence from Wikipedia]}} [INFER] {\textcolor{blue}{[Inference$_1$ from other samples]}} / {\textcolor{red}{[A sentence from Wikipedia]}}\\
    \midrule
    \textbf{Pattern} & {\textcolor{blue}{[BECAUSE]}} [statement$_1$] {\textcolor{blue}{[AND]}} [statement$_2$] {\textcolor{blue}{[INFER]}} [Inference$_1$] & {\textcolor{blue}{[A word from BECAUSE, AND, OR, INFER]}} / {\textcolor{red}{[A random word]}} [statement$_1$] {\textcolor{blue}{[A word from BECAUSE, AND, OR, INFER]}} / {\textcolor{red}{[A random word]}} [statement$_2$] {\textcolor{blue}{[A word from BECAUSE, AND, OR, INFER]}} / {\textcolor{red}{[A random word]}} [Inference$_1$]\\
    \midrule
    \textbf{Definition} & The definition of logical AND is as follows: {\textcolor{blue}{[The definition of AND from Wikipedia]}}. The definition of logical OR is as follows: {\textcolor{blue}{[The definition of OR from Wikipedia]}}. Based on the definitions, answer the following question. & The concepts of logical AND and logical OR have now been swapped. The definition of logical AND is as follows: {\textcolor{blue}{[The definition of OR from Wikipedia]}}. The definition of logical OR is as follows: {\textcolor{blue}{[The definition of AND from Wikipedia]}}. Based on the revised definitions, answer the following question.\\
    \bottomrule
  \end{tabularx}
  \end{small}
\caption{The comparison between raw data and data after replacement or modification operation from Entailment Bank. In-domain replace are printed in blue, and out-of-domain replace are printed in red.}
\label{tab:task}
\end{table*}

\noindent\textbf{Replacement:} Replacement for the \emph{Text}, \emph{Reasoning Chain} and \emph{Pattern}. This operation involves replacing the current content with content from another example within the same domain (in-domain) or with unrelated text (out-of-domain).
Through replace operation, we can observe which parts of the data are more important for establishing the logical reasoning of the large model. Furthermore, we can explore the model's robustness to disturbances and its ability to understand patterns.

\noindent\textbf{Modification:} To test the large model's understanding of logical rules, modifications are made to the definitions of logical concepts. For example, we modify the definitions of \emph{AND} and \emph{OR}. 
We follow the input examples with a statement that reassigns the original meaning of \emph{AND} to \emph{OR}, and vice versa. Given that the input examples utilize the standard interpretations of \emph{AND} and \emph{OR}, altering their definitions should result in an inversion of the corresponding relational statements in the output. If the model predominantly learns through probabilistic associations between tokens, the probability of correctly interchanging \emph{AND} and \emph{OR} in its output is expected to be low. However, if the model genuinely comprehends the logical symbols and their governing rules, it should accurately replace \emph{AND} with \emph{OR}, and \emph{OR} with \emph{AND} in the output, reflecting this new understanding.\footnote{For specific examples, please refer to Table~\ref{tab:task}.}

\section{Experiment}

In this section, we conduct extensive experiments to explore LLMs' ability for logic understanding.

\subsection{Models}
In exploring LLMs' ability to understand rules, we have employed two model series from the Open LLM Leaderboard\footnote{\url{https://huggingface.co/spaces/HuggingFaceH4/ open_llm_leaderboard}}, each with varying scales of parameter sizes, to conduct our experiments.
LLaMA2~\cite{llama2}, open-sourced and developed by Meta, represents a suite of pre-trained and fine-tuned LLMs. These models vary in complexity, featuring sizes from 7B to 70B parameters.
Additionally, we employed models from the Qwen series
\footnote{Qwen models from 7B to 72B are downloaded from \url{https://github.com/QwenLM/Qwen}. The outputs of the 200B model are obtained via API calls.}, which range in size from 7B to 200B parameters. These models have undergone stable pre-training on up to 3 trillion tokens of multilingual data, encompassing a broad spectrum of domains and languages with an emphasis on Chinese and English.
Among these, the 200B-parameter model is essentially the largest in terms of the number of parameters available to us.\footnote{We do not train the models; instead, we test these models on their in-context learning capabilities and abilities to understand logical rules through specific inputs.}

\subsection{Datasets}
As our experiments require intermediate reasoning steps, we utilized the dataset released by~\citealp{logicalcot}, known as LogicalCOT.\footnote{\url{https://huggingface.co/datasets/csitfun/LogiCoT}} The specific tasks include the following three types:

\noindent\textbf{Folio (Language to Logic):} This process involves translating natural language into a more formal logical notation, a fundamental task that requires comprehending and interpreting logical statements articulated in natural language and transforming them into a formalized logical framework.

\noindent\textbf{Entailment Bank (Inference Chains):} This instructional approach advances logical reasoning by requiring the model to ascertain the probability of a potential inference from a given set of premises. Subsequently, the model must delineate the sequence of logical deductions leading to the conclusion. Such an approach fosters deeper logical analysis and the capability to formulate cogent arguments. The examples provided for practice are formulated either in a symbolic language or articulated in natural language for greater accessibility and comprehension.

\noindent\textbf{MRC:} Machine Reading Comprehension (MRC) serves as the primary task for evaluating the reasoning capabilities of LLMs, wherein a model is provided with a passage and a corresponding question and is tasked with identifying the correct answer. This domain encompasses tasks that necessitate a deep comprehension of the provided text, often requiring the model to recognize, extract, or deduce information from the text. Models may be tasked with resolving scenarios depicted in the text, identifying fallacies within an argument, or determining information that could bolster or undermine a presented argument.

\begin{table*}[]
\small
\label{table_open_domain}
\center
\begin{tabular}{lccccccccccccc}
\toprule
\multirow{3}{*}{\textbf{Models}} & \multirow{3}{*}{\textbf{w/o}}              & \multicolumn{6}{c}{\textbf{4 In-context Examples}}                                                                                                & \multicolumn{6}{c}{\textbf{8 In-context Examples}}\\  
                                                         &                    &   \textbf{Raw}     & \multicolumn{2}{c}{\textbf{Text}}                    & \multicolumn{2}{c}{\textbf{Chain}}   &\textbf{Pattern}             & \textbf{Raw}      & \multicolumn{2}{c}{\textbf{Text}}                    & \multicolumn{2}{c}{\textbf{Chain}}           &\textbf{Pattern} \\ 
                                                        &                     &                   & $In$                 & $Out$                          & $In$                 & $Out$                   & $Random$          &                     & $In$                 & $Out$                         & $In$                 & $Out$                   & $Random$              \\ \midrule
\multicolumn{12}{c}{Entailment Bank}                                                                                                                                                                                                                                                                \\ \midrule
LLaMA2-7B-Chat                                          & 46.2              & 56.4                  & 42.2               & 49.0                            & 45.5             & 49.9                       & 53.8               & 57.1                  & 41.8               & 48.5                            & 46.7             & 48.2                       & 53.3      \\
LLaMA2-13B-Chat                                         & 72.2              & 76.7                  & 42.4               & 47.8                            & 44.5             & 65.2                       & 71.9               & 75.8                  & 45.4               & 45.4                            & 42.9             & 60.2                       & 73.0         \\ 
LLaMA2-70B-Chat                                         & 74.8              & 83.9                  & 82.3               & 80.8                            & 83.2             & 81.3                       & 83.8               & 84.1                  & 83.6               & 83.7                            & 82.5             & 81.8                       & 84.2  \\ 
Qwen-7B-Chat                                            & 53.5              & 62.7                  & 45.6               & 52.1                            & 48.8             & 50.8                       & 59.3               & 64.4                  & 43.3               & 44.1                            & 47.4             & 49.7                       & 60.5         \\ 
Qwen-14B-Chat                                           & 72.1              & 78.7                  & 50.9               & 52.6                            & 45.1             & 63.2                       & 73.5               & 76.6                  & 46.1               & 45.8                            & 48.7             & 62.3                       & 75.4        \\ 
Qwen-72B-Chat                                           & 76.4              & 85.9                  & 84.3               & 84.8                            & 85.6             & 85.0                       & 86.2               & 87.7                  & 86.1               & 86.5                            & 87.0             & 85.4                       & 86.6             \\ 
Qwen-200B-Chat                                          & 80.9              & 92.8                  & 90.2               & 91.8                            & 92.2             & 90.0                       & 93.4               & 92.6                  & 90.4               & 91.5                            & 92.3             & 88.8                       & 93.3          \\ \midrule
\multicolumn{12}{c}{Folio}                                                                                                                                                                                                                                                      \\ \midrule
LLaMA2-7B-Chat                                          & 45.4              & 57.9                  & 40.2               & 41.8                            & /                & /                          & 55.0               & 60.2                  & 38.7               & 39.9                            & /                & /                          & 55.6   \\
LLaMA2-13B-Chat                                         & 68.2              & 72.4                  & 45.8               & 45.1                            & /                & /                          & 63.7               & 72.5                  & 44.1               & 43.4                            & /                & /                          & 64.5     \\ 
LLaMA2-70B-Chat                                         & 73.8              & 82.6                  & 80.4               & 80.5                            & /                & /                          & 82.7               & 83.0                  & 79.4               & 80.9                            & /                & /                          & 82.6    \\ 
Qwen-7B-Chat                                            & 60.2              & 68.6                  & 46.8               & 46.2                            & /                & /                          & 68.9               & 69.0                  & 48.9               & 49.2                            & /                & /                          & 68.6      \\ 
Qwen-14B-Chat                                           & 72.8              & 84.6                  & 63.2               & 65.8                            & /                & /                          & 83.4               & 85.1                  & 62.4               & 63.8                            & /                & /                          & 83.9        \\ 
Qwen-72B-Chat                                           & 84.6              & 93.7                  & 90.2               & 92.2                            & /                & /                          & 94.6               & 92.9                  & 90.4               & 91.5                            & /                & /                          & 91.0         \\ 
Qwen-200B-Chat                                          & 85.8              & 94.2                  & 92.5               & 94.0                            & /                & /                          & 95.1               & 93.9                  & 91.3               & 93.8                            & /                & /                          & 93.5           \\ \midrule
\multicolumn{12}{c}{MRC}                                                                                                                                                                                                                                                      \\ \midrule
LLaMA2-7B-Chat                                          & 30.8              & 32.1                  & 27.6               & 28.7                            & 28.1             & 27.6                       & /                  & 33.2                  & 27.7               & 28.5                            & 27.6             & 28.0                       & /         \\
LLaMA2-13B-Chat                                         & 40.2              & 42.0                  & 36.2               & 38.7                            & 35.1             & 36.6                       & /                  & 45.2                  & 38.1               & 40.3                            & 40.4             & 40.7                       & /             \\ 
LLaMA2-70B-Chat                                         & 59.2              & 65.5                  & 62.0               & 62.6                            & 63.1             & 62.9                       & /                  & 67.8                  & 64.1               & 64.7                            & 46.7             & 48.2                       & /        \\ 
Qwen-7B-Chat                                            & 43.4              & 56.6                  & 53.3               & 53.7                            & 54.0             & 54.9                       & /                  & 60.4                  & 58.2               & 58.0                            & 65.2             & 65.8                       & /         \\ 
Qwen-14B-Chat                                           & 60.5              & 68.9                  & 61.3               & 62.8                            & 63.4             & 63.2                       & /                  & 69.2                  & 62.4               & 63.1                            & 64.0             & 64.5                       & /            \\ 
Qwen-72B-Chat                                           & 74.6              & 79.5                  & 78.1               & 78.8                            & 80.1             & 78.2                       & /                  & 80.0                  & 78.4               & 79.3                            & 80.4             & 78.5                       & /        \\ 
Qwen-200B-Chat                                          & 78.9              & 80.6                  & 80.2               & 80.1                            & 79.3             & 79.1                       & /                  & 81.9                  & 80.5               & 79.7                            & 79.0             & 80.1                      & /              \\ \midrule
\end{tabular}
\caption{Results for the LLaMA and Qwen model series on the logical datasets. ($Acc.$ \%) Here, \textbf{w/o} stands for \emph{without in-context example}, while \textbf{Raw} denotes results enhanced \emph{with regular in-context examples}.}
\label{table_res_main}
\end{table*}

\noindent\textbf{Data Source for Replacement:}
We utilize other samples as the in-domain data.
For out-of-domain data, we use the English Wikipedia (2020/03/01) \footnote{\url{https://dumps.wikimedia.org/enwiki/}} as the out-of-domain data source.
We randomly selected a paragraph from one of the 2.6 billion documents to replace the content of the text and reasoning chain.

\subsection{Influence of In-Context Examples}
In Table~\ref{table_res_main}, we observe a positive correlation between the number of in-context examples and the accuracy of the model's predictions. The improvement brought about by using in-context examples is quite evident, which is consistent with ~\citealp{nlp_ins, meta_icl, cross_task_ins}. However, in our results, using 8 examples does not yield a significant enhancement over using 4 examples. 
Furthermore, this relationship is amplified as model size scales (from 7B to 200B parameters), suggesting that larger models benefit disproportionately from an increased number of examples. Additionally, in-context examples contribute to the standardization of the output format, thereby facilitating the generation of outputs that are consistent with the expected structure.

\begin{figure}[t]
\centering
\includegraphics[width=0.9\columnwidth]{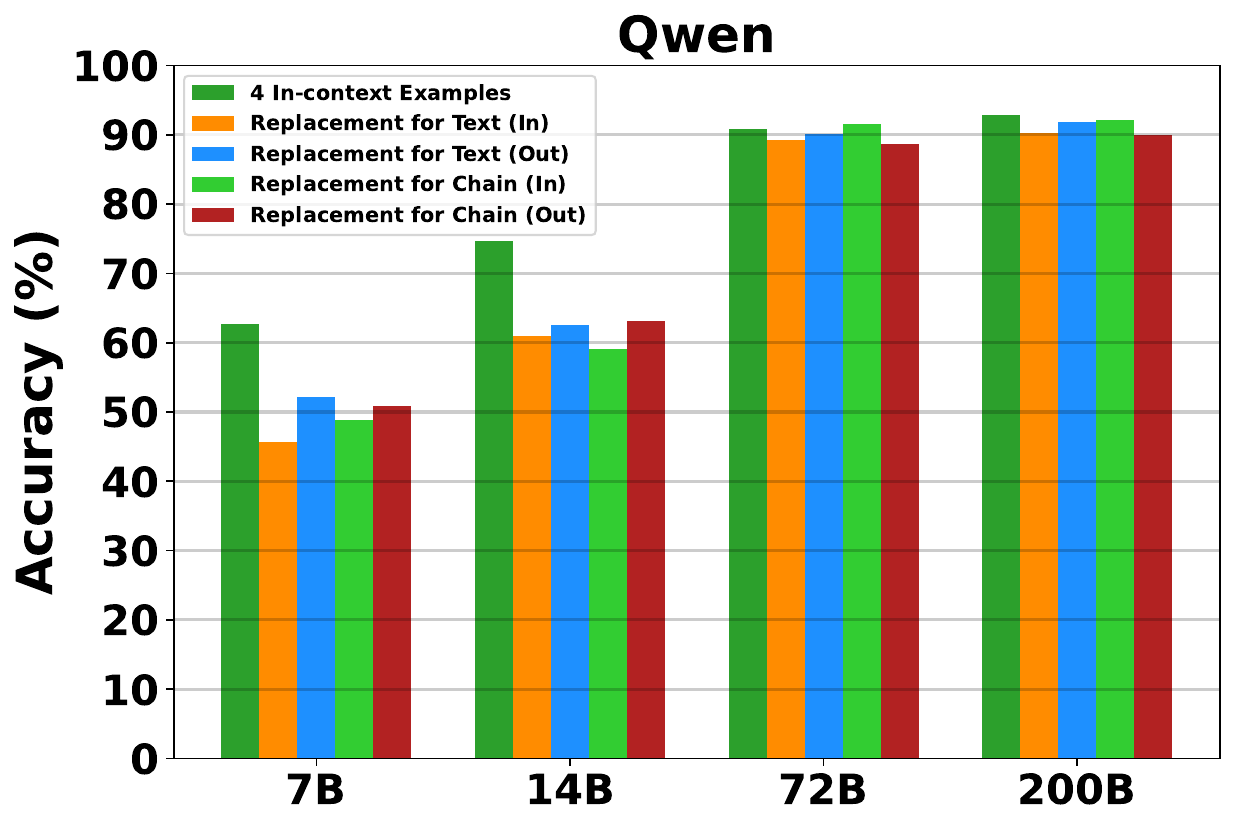}
\caption{The impact of different replacement parts on Entailment Bank for Qwen series models' performance.}
\label{fig_eb_qwen}
\end{figure}

\begin{figure}[t]
\centering
\includegraphics[width=0.9\columnwidth]{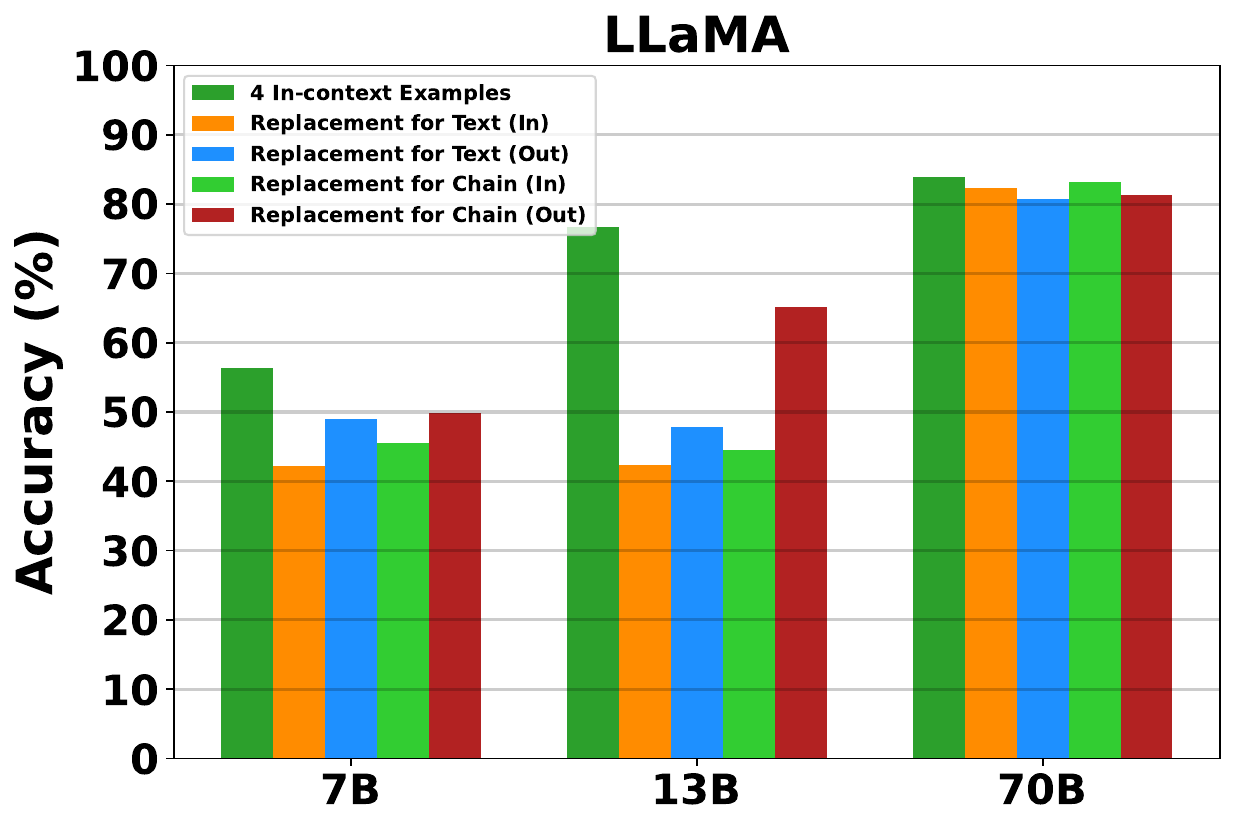}
\caption{The impact of different replacement parts on Entailment Bank for LLaMA series models' performance.}
\label{fig_eb_llama}
\end{figure}

\subsection{Influence of Texts} 

\noindent\textbf{In-Domain:} We observe that smaller-scale models (7B/13B) exhibit a pronounced decline in accuracy when the context provided in examples is modified, as delineated in Table~\ref{table_res_main}. Conversely, as we can see from Figure~\ref{fig_eb_qwen} and Figure~\ref{fig_eb_llama}, larger models (70B/200B) demonstrate resilience to such contextual manipulations, with negligible impacts on accuracy.
We hypothesize that the augmented capacity of larger models equips them with enhanced resistance to perturbations of textual input, enabling them to extract and retain salient information from a prescribed format while remaining focused on the central question. In contrast, smaller models appear to be more susceptible to textual interference, predominantly assimilating linguistic details from the context, which consequently precipitates inaccuracies in addressing the question.

\noindent\textbf{Out-of-Domain:}
When utilizing out-of-domain data, the observations bear a resemblance to those gleaned from in-domain data. However, a clear disparity emerges in the robustness of smaller models compared to their larger counterparts when confronted with out-of-domain text. Smaller models exhibit a marked decrease in performance. In contrast, the performance of larger models remains largely stable, showing a negligible impact from such perturbations.

Paradoxically, when examining performance on in-domain text, we find that models trained with out-of-domain data not only match but occasionally surpass the outcomes attained with in-domain data. This finding runs counter to conventional expectations. The question arises as to why models yield superior results when trained on seemingly irrelevant data and why this enhancement is more pronounced in smaller models.
We hypothesize that the enhanced performance can be attributed to the greater divergence of out-of-domain data from the original data distribution. Such divergence may enable the model to distinguish irrelevant text with heightened clarity, thereby sharpening its focus on content pertinent to the task at hand.

\begin{figure*}[htb]
  \centering
  \begin{minipage}[b]{\linewidth}
    \includegraphics[width=.95\linewidth]{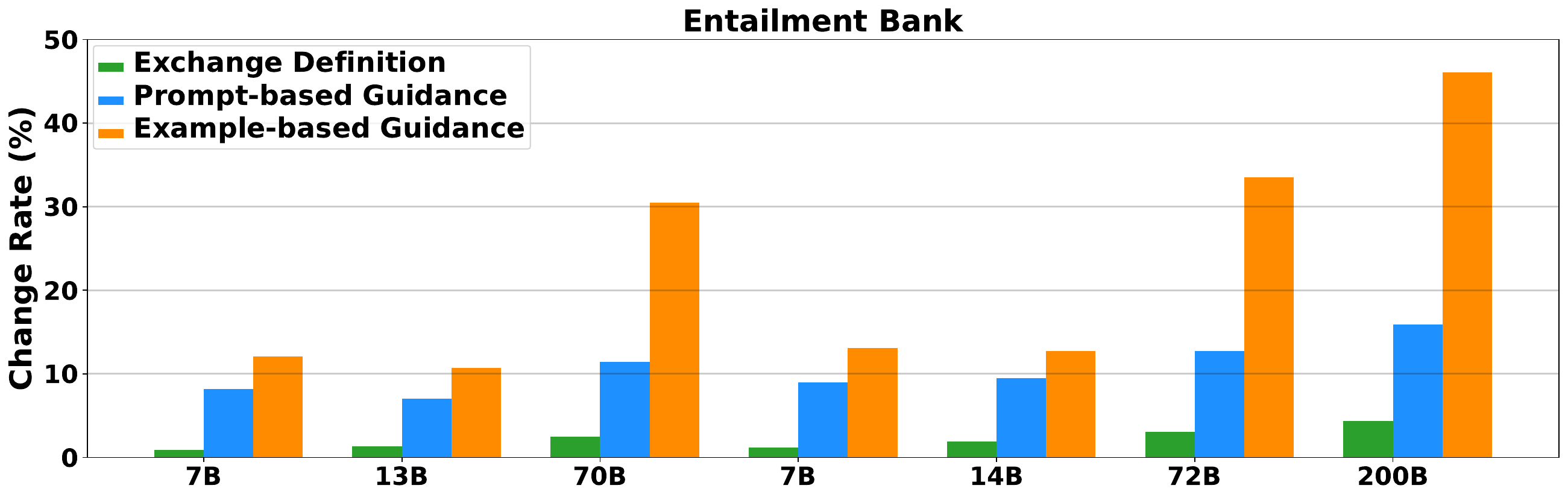}
    \label{fig:sub1}
  \end{minipage}
  \hfill 
  \begin{minipage}[b]{\linewidth}
    \includegraphics[width=.95\linewidth]{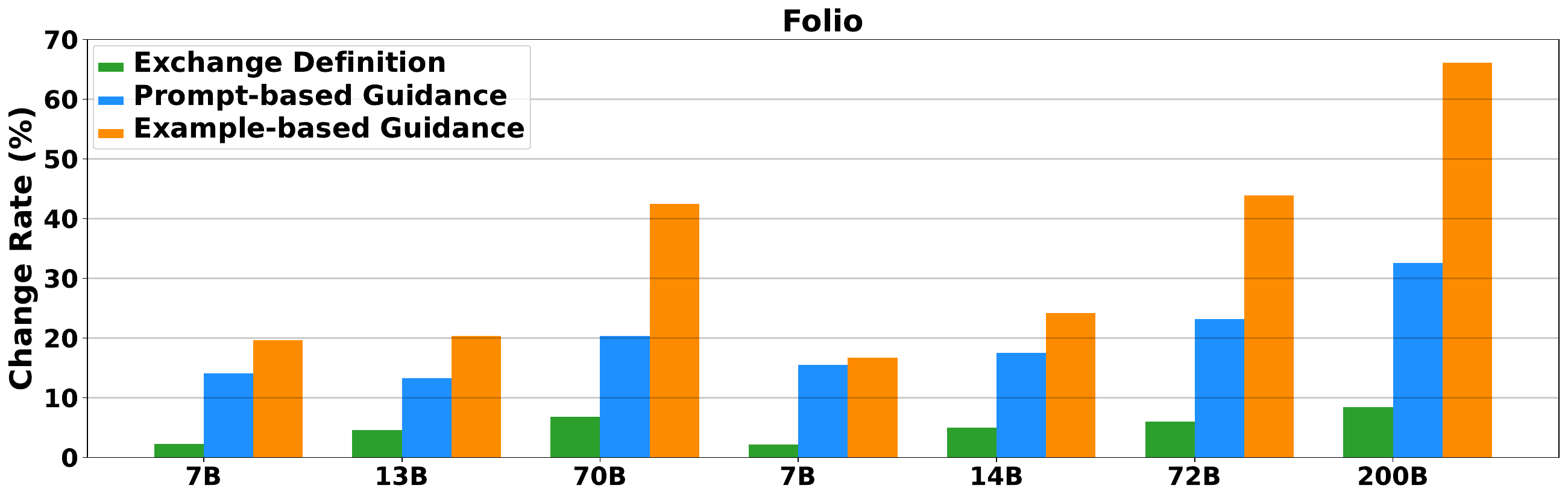}
    \label{fig:sub2}
  \end{minipage}
\vspace{-1em}
\caption{Results of different scales of LLaMA and Qwen models over Entailment Bank when using different settings. Each target example has 4 in-context samples as the demonstration.}

\end{figure*}

\subsection{Influence of Reasoning Chain}
\noindent\textbf{In-Domain:}
Upon replacing the reasoning chains in our experiment, we observed phenomena analogous to those documented during text substitution. Notably, smaller models demonstrated a disproportionately substantial decline in accuracy, with a large reduction for 7B and 13B models as opposed to a slight decrease for 70B and 200B LLM.\footnote{For details, see Table~\ref{table_res_main}.} Regardless of the model size, the decrease in accuracy of reasoning outcomes, engendered by the substitution of reasoning chains, proved less pronounced than that occasioned by text replacement. This disparity can be attributed to the text's integral role in defining the problem and potential solutions, which facilitates the model's ability to forge connections between the input and the expected output, thereby mitigating the influence of alterations in the reasoning chain.

\noindent\textbf{Out-of-Domain:}
Upon substituting out-of-domain data for the reasoning chain, we observed an unexpected phenomenon. It can be seen from Figure~\ref{fig_eb_qwen} and Figure~\ref{fig_eb_llama} that 7B and 13B models exhibited only a modest reduction in reasoning performance when utilizing out-of-domain data in Entailment Bank, as opposed to a more substantial decline with in-domain data. Conversely, 70B and 200B models demonstrated a more pronounced decrease in performance with out-of-domain data compared to in-domain data. This divergence in behavior between smaller and larger models warrants further investigation.
We hypothesize that the stark contrast in data distributions between out-of-domain and original datasets prompts smaller models to disregard the textual content within the reasoning chains. Consequently, these models form a direct association between the input text and the corresponding output answer, largely ignoring intermediate reasoning steps. In contrast, larger models, equipped with more robust comprehension capabilities, are significantly affected by the content of the reasoning chains. This heightened sensitivity to the reasoning process results in more substantial disruptions in their output when confronted with out-of-domain data.

\subsection{Influence of Pattern}
Our investigation extends to the model's sensitivity to substitutions of specific patterns within the text. We conducted experiments where lexical items such as \emph{[AND]}, \emph{[OR]}, and \emph{[BECAUSE]} were interchanged (e.g., \emph{[AND]} $\leftrightarrow$ \emph{[OR]}, \emph{[OR]} $\leftrightarrow$ \emph{[BECAUSE]}). Notably, substituting \emph{[AND]} for \emph{[OR]} resulted in the model producing outputs where the corresponding terms were interchanged. However, it can be seen from Table~\ref{table_res_main} that despite maintaining the logical relationships among these conditions, such alterations did not significantly impact the model's output accuracy. Additionally, introducing non-sequitur substitutions (e.g., \emph{[AND]} $\leftrightarrow$ \emph{[APPLE/BANANA]}) did not meaningfully reduce the accuracy of the model's outputs.

These findings suggest that the model primarily recognizes the necessity for a syntactic linkage between adjoining sentences, as signified by the presence of markers enclosed within brackets \emph{[]}, rather than comprehending the nuanced semantic influence exerted by logical connectives such as \emph{[AND]} or \emph{[OR]}. The implication is that the model may be relying on surface-level cues to maintain coherence rather than deeply processing the logical relationships underpinning the text's structure.

\subsection{Test for Logical Understanding Ability}
To evaluate the model's grasp of logical reasoning, we implemented a methodology that introduces prompts subsequent to the examples. This approach serves to ascertain the model's comprehension of logical constructs.

\noindent\textbf{Modify Symbols and Logical Predicates:}
It has been observed that altering symbols and logical predicates within a given context does not compromise the performance of large language models in terms of generating output. However, these outputs are logically inconsistent at a relational level. For instance, conclusions predicated on the use of an [AND] logical connector do not retain their validity when the [OR] connector is substituted.

\noindent\textbf{Modification of Logical Predicates:}
Our approach utilizes the definitions of logical predicates and symbols as delineated by Wikipedia. We introduce a prompt subsequent to an in-context example.\footnote{For deatails, see Table~\ref{tab:task}}. This is done to evaluate the model's comprehension of logical terminology—[AND], [OR], and others. The expectation is that the model will generate text where conjunctions previously denoted by [AND] ([OR]) are now conveyed through [OR] ([AND]), with a higher rate of modification indicating a better result. Examination of the data reveals that smaller models (7B/13B) demonstrate a negligible modification rate below 1\%, while the modification rate for larger models is below 5\%. This suggests that, although the smaller models seem to address logic-related queries adequately, their grasp of logical semantics in particular scenarios is limited. Similarly, the performance of the larger models (70B/200B) is suboptimal. They exhibit a rudimentary understanding of these logical predicates—presumably acquired during their pre-training phase—but fall short of achieving satisfactory performance.

It is worth noting that in such scenarios, larger models may produce outputs that reveal underlying confusions or rationales. Here is an example output by Qwen-200B-Chat: \emph{I apologize, but there seems to be a misunderstanding. The provided examples don't adhere to the new definitions of logical AND and OR. However, based on the modified meanings of logical OR (where both conditions must be true for the conclusion to hold), we can infer that ...}

\subsection{Enhancing Logical Comprehension Ability for LLM}

The question arises whether it is possible to augment the logical reasoning capabilities of large-scale models without resorting to further training. To address this, we have explored two distinct approaches:

\noindent\textbf{Prompt-based Guidance: }
Expanding upon the modified definitions, this study incorporated a supplementary instructional prompt directing the model to interchange the logical operators [OR] with [AND], and [AND] with [OR], while ensuring grammatical correctness and logical consistency. Subsequent to the application of this prompt, a discernible enhancement in the model's performance in executing operator swaps was observed; however, the improvements did not fulfill our expectations.

\noindent\textbf{Example-based Guidance:}
The capacity for comprehension enhancement through mere prompt-based instruction in models is constrained. To address this, we endeavored to enrich the instructional framework by supplementing guiding prompts with illustrative modifications. For example, we provided a practice scenario as follows: "Original Statement: '[BECAUSE] [Statement$_1$] [AND] [Statement$_2$] [INFER] [Inference$_1$].' Your Modification: '[BECAUSE] [Statement$_1$] [OR] [Statement$_2$] [INFER] [Inference$_1$].' Now, it is your turn to modify." Subsequent to the implementation of both guiding prompts and contextualized example-based instruction, there was an observable augmentation in the modification rate by the large-scale model to over 40~50\%. This increment indicates a substantial dependency of the model on contextually provided examples. The recurrence of certain logical operator predicate patterns in precedent examples suggests that mere reliance on definitions or prompts is inadequate for mitigating these patterns. Instead, incorporating examples that mirror the anticipated format of modifications is imperative for realizing a significant improvement. Thus, the exploration of methods to enhance the model's logical reasoning capabilities independent of context-based examples constitutes an avenue for future research.

\section{Conclusion}
In this study, we investigate the capacity of LLMs, with parameters varying from 7B to 200B, to comprehend logical rules. The observed performance disparity between smaller and larger models indicates that size alone does not guarantee a profound understanding of logical constructs. While larger models may show traces of semantic learning, their outputs often lack logical validity when faced with swapped logical predicates. Our findings suggest that while LLMs may improve their logical reasoning performance through in-context learning and methodologies such as COT, these enhancements do not equate to a genuine understanding of logical operations and definitions, nor do they necessarily confer the capability for logical reasoning.

\section*{Limitations}
Despite employing prompts and in-context examples that ostensibly improve the model's capacity for logical reasoning, the enhancement remains marginal. To date, a method that markedly augments the model's comprehension through in-context learning has not been identified. The prevailing pre-training mechanism focuses on next-token prediction by estimating the subsequent word based on a probability distribution and may not be ideally suited for logical tasks. These tasks often necessitate the processing of longer-span dependencies and the integration of global information for effective reasoning. Consequently, we believe that devising an alternative pre-training strategy tailored to these requirements presents a promising avenue for future research.

\bibliography{anthology,custom}

\begin{thebibliography}{32}
\expandafter\ifx\csname natexlab\endcsname\relax\def\natexlab#1{#1}\fi

\bibitem[{Biderman et~al.(2023)Biderman, Schoelkopf, Anthony, Bradley, O'Brien, Hallahan, Khan, Purohit, Prashanth, Raff, Skowron, Sutawika, and van~der Wal}]{DBLP:conf/icml/BidermanSABOHKP23}
Stella Biderman, Hailey Schoelkopf, Quentin~Gregory Anthony, Herbie Bradley, Kyle O'Brien, Eric Hallahan, Mohammad~Aflah Khan, Shivanshu Purohit, USVSN~Sai Prashanth, Edward Raff, Aviya Skowron, Lintang Sutawika, and Oskar van~der Wal. 2023.
\newblock Pythia: {A} suite for analyzing large language models across training and scaling.
\newblock In \emph{ICML}, volume 202 of \emph{Proceedings of Machine Learning Research}, pages 2397--2430. {PMLR}.

\bibitem[{Bommasani et~al.(2021)Bommasani, Hudson, Adeli, Altman, Arora, von Arx, Bernstein, Bohg, Bosselut, Brunskill, Brynjolfsson, Buch, Card, Castellon, Chatterji, Chen, Creel, Davis, Demszky, Donahue, Doumbouya, Durmus, Ermon, Etchemendy, Ethayarajh, Fei{-}Fei, Finn, Gale, Gillespie, Goel, Goodman, Grossman, Guha, Hashimoto, Henderson, Hewitt, Ho, Hong, Hsu, Huang, Icard, Jain, Jurafsky, Kalluri, Karamcheti, Keeling, Khani, Khattab, Koh, Krass, Krishna, Kuditipudi, and et~al.}]{opp_of_llm}
Rishi Bommasani, Drew~A. Hudson, Ehsan Adeli, Russ~B. Altman, Simran Arora, Sydney von Arx, Michael~S. Bernstein, Jeannette Bohg, Antoine Bosselut, Emma Brunskill, Erik Brynjolfsson, Shyamal Buch, Dallas Card, Rodrigo Castellon, Niladri~S. Chatterji, Annie~S. Chen, Kathleen Creel, Jared~Quincy Davis, Dorottya Demszky, Chris Donahue, Moussa Doumbouya, Esin Durmus, Stefano Ermon, John Etchemendy, Kawin Ethayarajh, Li~Fei{-}Fei, Chelsea Finn, Trevor Gale, Lauren Gillespie, Karan Goel, Noah~D. Goodman, Shelby Grossman, Neel Guha, Tatsunori Hashimoto, Peter Henderson, John Hewitt, Daniel~E. Ho, Jenny Hong, Kyle Hsu, Jing Huang, Thomas Icard, Saahil Jain, Dan Jurafsky, Pratyusha Kalluri, Siddharth Karamcheti, Geoff Keeling, Fereshte Khani, Omar Khattab, Pang~Wei Koh, Mark~S. Krass, Ranjay Krishna, Rohith Kuditipudi, and et~al. 2021.
\newblock \href {http://arxiv.org/abs/2108.07258} {On the opportunities and risks of foundation models}.
\newblock \emph{CoRR}, abs/2108.07258.

\bibitem[{Brown et~al.(2020{\natexlab{a}})Brown, Mann, Ryder, Subbiah, Kaplan, Dhariwal, Neelakantan, Shyam, Sastry, Askell, Agarwal, Herbert{-}Voss, Krueger, Henighan, Child, Ramesh, Ziegler, Wu, Winter, Hesse, Chen, Sigler, Litwin, Gray, Chess, Clark, Berner, McCandlish, Radford, Sutskever, and Amodei}]{gpt3}
Tom~B. Brown, Benjamin Mann, Nick Ryder, Melanie Subbiah, Jared Kaplan, Prafulla Dhariwal, Arvind Neelakantan, Pranav Shyam, Girish Sastry, Amanda Askell, Sandhini Agarwal, Ariel Herbert{-}Voss, Gretchen Krueger, Tom Henighan, Rewon Child, Aditya Ramesh, Daniel~M. Ziegler, Jeffrey Wu, Clemens Winter, Christopher Hesse, Mark Chen, Eric Sigler, Mateusz Litwin, Scott Gray, Benjamin Chess, Jack Clark, Christopher Berner, Sam McCandlish, Alec Radford, Ilya Sutskever, and Dario Amodei. 2020{\natexlab{a}}.
\newblock \href {https://proceedings.neurips.cc/paper/2020/hash/1457c0d6bfcb4967418bfb8ac142f64a-Abstract.html} {Language models are few-shot learners}.
\newblock In \emph{Advances in Neural Information Processing Systems 33: Annual Conference on Neural Information Processing Systems 2020, NeurIPS 2020, December 6-12, 2020, virtual}.

\bibitem[{Brown et~al.(2020{\natexlab{b}})Brown, Mann, Ryder, Subbiah, Kaplan, Dhariwal, Neelakantan, Shyam, Sastry, Askell, Agarwal, Herbert{-}Voss, Krueger, Henighan, Child, Ramesh, Ziegler, Wu, Winter, Hesse, Chen, Sigler, Litwin, Gray, Chess, Clark, Berner, McCandlish, Radford, Sutskever, and Amodei}]{DBLP:conf/nips/BrownMRSKDNSSAA20}
Tom~B. Brown, Benjamin Mann, Nick Ryder, Melanie Subbiah, Jared Kaplan, Prafulla Dhariwal, Arvind Neelakantan, Pranav Shyam, Girish Sastry, Amanda Askell, Sandhini Agarwal, Ariel Herbert{-}Voss, Gretchen Krueger, Tom Henighan, Rewon Child, Aditya Ramesh, Daniel~M. Ziegler, Jeffrey Wu, Clemens Winter, Christopher Hesse, Mark Chen, Eric Sigler, Mateusz Litwin, Scott Gray, Benjamin Chess, Jack Clark, Christopher Berner, Sam McCandlish, Alec Radford, Ilya Sutskever, and Dario Amodei. 2020{\natexlab{b}}.
\newblock Language models are few-shot learners.
\newblock In \emph{NeurIPS}.

\bibitem[{Chen et~al.(2022)Chen, Zhong, Zha, Karypis, and He}]{meta_icl}
Yanda Chen, Ruiqi Zhong, Sheng Zha, George Karypis, and He~He. 2022.
\newblock \href {https://doi.org/10.18653/V1/2022.ACL-LONG.53} {Meta-learning via language model in-context tuning}.
\newblock In \emph{Proceedings of the 60th Annual Meeting of the Association for Computational Linguistics (Volume 1: Long Papers), {ACL} 2022, Dublin, Ireland, May 22-27, 2022}, pages 719--730. Association for Computational Linguistics.

\bibitem[{Gokhale et~al.(2021)Gokhale, Chaudhary, Banerjee, Baral, and Yang}]{semantically_robust_optimization}
Tejas Gokhale, Abhishek Chaudhary, Pratyay Banerjee, Chitta Baral, and Yezhou Yang. 2021.
\newblock Semantically distributed robust optimization for vision-and-language inference.
\newblock \emph{arXiv preprint arXiv:2110.07165}.

\bibitem[{Kaushik et~al.(2019)Kaushik, Hovy, and Lipton}]{kaushik2019learning}
Divyansh Kaushik, Eduard Hovy, and Zachary Lipton. 2019.
\newblock Learning the difference that makes a difference with counterfactually-augmented data.
\newblock In \emph{International Conference on Learning Representations}.

\bibitem[{Li and Srikumar(2019)}]{li-srikumar-2019-augmenting}
Tao Li and Vivek Srikumar. 2019.
\newblock \href {https://doi.org/10.18653/v1/P19-1028} {Augmenting neural networks with first-order logic}.
\newblock In \emph{Proceedings of the 57th Annual Meeting of the Association for Computational Linguistics}, pages 292--302, Florence, Italy. Association for Computational Linguistics.

\bibitem[{Liu et~al.(2023)Liu, Teng, Cui, Zhang, Zhou, and Zhang}]{logicalcot}
Hanmeng Liu, Zhiyang Teng, Leyang Cui, Chaoli Zhang, Qiji Zhou, and Yue Zhang. 2023.
\newblock \href {https://aclanthology.org/2023.findings-emnlp.191} {Logicot: Logical chain-of-thought instruction tuning}.
\newblock In \emph{Findings of the Association for Computational Linguistics: {EMNLP} 2023, Singapore, December 6-10, 2023}, pages 2908--2921. Association for Computational Linguistics.

\bibitem[{Lu et~al.(2021)Lu, Grover, Abbeel, and Mordatch}]{tf_as_eng}
Kevin Lu, Aditya Grover, Pieter Abbeel, and Igor Mordatch. 2021.
\newblock \href {http://arxiv.org/abs/2103.05247} {Pretrained transformers as universal computation engines}.
\newblock \emph{CoRR}, abs/2103.05247.

\bibitem[{MacCartney and Manning(2007)}]{maccartney-manning-2007-natural}
Bill MacCartney and Christopher~D. Manning. 2007.
\newblock \href {https://aclanthology.org/W07-1431} {Natural logic for textual inference}.
\newblock In \emph{Proceedings of the {ACL}-{PASCAL} Workshop on Textual Entailment and Paraphrasing}, pages 193--200, Prague. Association for Computational Linguistics.

\bibitem[{Min et~al.(2021)Min, Ross, Sulem, Veyseh, Nguyen, Sainz, Agirre, Heintz, and Roth}]{DBLP:journals/corr/abs-2111-01243}
Bonan Min, Hayley Ross, Elior Sulem, Amir Pouran~Ben Veyseh, Thien~Huu Nguyen, Oscar Sainz, Eneko Agirre, Ilana Heintz, and Dan Roth. 2021.
\newblock Recent advances in natural language processing via large pre-trained language models: {A} survey.
\newblock \emph{CoRR}, abs/2111.01243.

\bibitem[{Min et~al.(2022)Min, Lyu, Holtzman, Artetxe, Lewis, Hajishirzi, and Zettlemoyer}]{rethink_in_context}
Sewon Min, Xinxi Lyu, Ari Holtzman, Mikel Artetxe, Mike Lewis, Hannaneh Hajishirzi, and Luke Zettlemoyer. 2022.
\newblock \href {https://doi.org/10.18653/V1/2022.EMNLP-MAIN.759} {Rethinking the role of demonstrations: What makes in-context learning work?}
\newblock In \emph{Proceedings of the 2022 Conference on Empirical Methods in Natural Language Processing, {EMNLP} 2022, Abu Dhabi, United Arab Emirates, December 7-11, 2022}, pages 11048--11064. Association for Computational Linguistics.

\bibitem[{Mishra et~al.(2022{\natexlab{a}})Mishra, Khashabi, Baral, and Hajishirzi}]{nlp_ins}
Swaroop Mishra, Daniel Khashabi, Chitta Baral, and Hannaneh Hajishirzi. 2022{\natexlab{a}}.
\newblock \href {https://doi.org/10.18653/V1/2022.ACL-LONG.244} {Cross-task generalization via natural language crowdsourcing instructions}.
\newblock In \emph{Proceedings of the 60th Annual Meeting of the Association for Computational Linguistics (Volume 1: Long Papers), {ACL} 2022, Dublin, Ireland, May 22-27, 2022}, pages 3470--3487. Association for Computational Linguistics.

\bibitem[{Mishra et~al.(2022{\natexlab{b}})Mishra, Khashabi, Baral, and Hajishirzi}]{cross_task_ins}
Swaroop Mishra, Daniel Khashabi, Chitta Baral, and Hannaneh Hajishirzi. 2022{\natexlab{b}}.
\newblock \href {https://doi.org/10.18653/V1/2022.ACL-LONG.244} {Cross-task generalization via natural language crowdsourcing instructions}.
\newblock In \emph{Proceedings of the 60th Annual Meeting of the Association for Computational Linguistics (Volume 1: Long Papers), {ACL} 2022, Dublin, Ireland, May 22-27, 2022}, pages 3470--3487. Association for Computational Linguistics.

\bibitem[{Muennighoff et~al.(2023)Muennighoff, Wang, Sutawika, Roberts, Biderman, Scao, Bari, Shen, Yong, Schoelkopf, Tang, Radev, Aji, Almubarak, Albanie, Alyafeai, Webson, Raff, and Raffel}]{DBLP:conf/acl/MuennighoffWSRB23}
Niklas Muennighoff, Thomas Wang, Lintang Sutawika, Adam Roberts, Stella Biderman, Teven~Le Scao, M.~Saiful Bari, Sheng Shen, Zheng~Xin Yong, Hailey Schoelkopf, Xiangru Tang, Dragomir Radev, Alham~Fikri Aji, Khalid Almubarak, Samuel Albanie, Zaid Alyafeai, Albert Webson, Edward Raff, and Colin Raffel. 2023.
\newblock Crosslingual generalization through multitask finetuning.
\newblock In \emph{ACL}, pages 15991--16111. Association for Computational Linguistics.

\bibitem[{Ouyang et~al.(2022{\natexlab{a}})Ouyang, Wu, Jiang, Almeida, Wainwright, Mishkin, Zhang, Agarwal, Slama, Ray, Schulman, Hilton, Kelton, Miller, Simens, Askell, Welinder, Christiano, Leike, and Lowe}]{instructGPT}
Long Ouyang, Jeffrey Wu, Xu~Jiang, Diogo Almeida, Carroll~L. Wainwright, Pamela Mishkin, Chong Zhang, Sandhini Agarwal, Katarina Slama, Alex Ray, John Schulman, Jacob Hilton, Fraser Kelton, Luke Miller, Maddie Simens, Amanda Askell, Peter Welinder, Paul~F. Christiano, Jan Leike, and Ryan Lowe. 2022{\natexlab{a}}.
\newblock \href {http://papers.nips.cc/paper\_files/paper/2022/hash/b1efde53be364a73914f58805a001731-Abstract-Conference.html} {Training language models to follow instructions with human feedback}.
\newblock In \emph{NeurIPS}.

\bibitem[{Ouyang et~al.(2022{\natexlab{b}})Ouyang, Wu, Jiang, Almeida, Wainwright, Mishkin, Zhang, Agarwal, Slama, Ray, Schulman, Hilton, Kelton, Miller, Simens, Askell, Welinder, Christiano, Leike, and Lowe}]{DBLP:conf/nips/Ouyang0JAWMZASR22}
Long Ouyang, Jeffrey Wu, Xu~Jiang, Diogo Almeida, Carroll~L. Wainwright, Pamela Mishkin, Chong Zhang, Sandhini Agarwal, Katarina Slama, Alex Ray, John Schulman, Jacob Hilton, Fraser Kelton, Luke Miller, Maddie Simens, Amanda Askell, Peter Welinder, Paul~F. Christiano, Jan Leike, and Ryan Lowe. 2022{\natexlab{b}}.
\newblock Training language models to follow instructions with human feedback.
\newblock In \emph{NeurIPS}.

\bibitem[{Qiu et~al.(2020)Qiu, Sun, Xu, Shao, Dai, and Huang}]{DBLP:journals/corr/abs-2003-08271}
Xipeng Qiu, Tianxiang Sun, Yige Xu, Yunfan Shao, Ning Dai, and Xuanjing Huang. 2020.
\newblock Pre-trained models for natural language processing: {A} survey.
\newblock \emph{CoRR}, abs/2003.08271.

\bibitem[{Scao et~al.(2022)Scao, Fan, Akiki, Pavlick, Ilic, Hesslow, Castagn{\'{e}}, Luccioni, Yvon, Gall{\'{e}}, Tow, Rush, Biderman, Webson, Ammanamanchi, Wang, Sagot, Muennighoff, del Moral, Ruwase, Bawden, Bekman, McMillan{-}Major, Beltagy, Nguyen, Saulnier, Tan, Suarez, Sanh, Lauren{\c{c}}on, Jernite, Launay, Mitchell, Raffel, Gokaslan, Simhi, Soroa, Aji, Alfassy, Rogers, Nitzav, Xu, Mou, Emezue, Klamm, Leong, van Strien, Adelani, and et~al.}]{DBLP:journals/corr/abs-2211-05100}
Teven~Le Scao, Angela Fan, Christopher Akiki, Ellie Pavlick, Suzana Ilic, Daniel Hesslow, Roman Castagn{\'{e}}, Alexandra~Sasha Luccioni, Fran{\c{c}}ois Yvon, Matthias Gall{\'{e}}, Jonathan Tow, Alexander~M. Rush, Stella Biderman, Albert Webson, Pawan~Sasanka Ammanamanchi, Thomas Wang, Beno{\^{\i}}t Sagot, Niklas Muennighoff, Albert~Villanova del Moral, Olatunji Ruwase, Rachel Bawden, Stas Bekman, Angelina McMillan{-}Major, Iz~Beltagy, Huu Nguyen, Lucile Saulnier, Samson Tan, Pedro~Ortiz Suarez, Victor Sanh, Hugo Lauren{\c{c}}on, Yacine Jernite, Julien Launay, Margaret Mitchell, Colin Raffel, Aaron Gokaslan, Adi Simhi, Aitor Soroa, Alham~Fikri Aji, Amit Alfassy, Anna Rogers, Ariel~Kreisberg Nitzav, Canwen Xu, Chenghao Mou, Chris Emezue, Christopher Klamm, Colin Leong, Daniel van Strien, David~Ifeoluwa Adelani, and et~al. 2022.
\newblock {BLOOM:} {A} 176b-parameter open-access multilingual language model.
\newblock \emph{CoRR}, abs/2211.05100.

\bibitem[{Taylor et~al.(2022)Taylor, Kardas, Cucurull, Scialom, Hartshorn, Saravia, Poulton, Kerkez, and Stojnic}]{DBLP:journals/corr/abs-2211-09085}
Ross Taylor, Marcin Kardas, Guillem Cucurull, Thomas Scialom, Anthony Hartshorn, Elvis Saravia, Andrew Poulton, Viktor Kerkez, and Robert Stojnic. 2022.
\newblock Galactica: {A} large language model for science.
\newblock \emph{CoRR}, abs/2211.09085.

\bibitem[{Touvron et~al.(2023{\natexlab{a}})Touvron, Lavril, Izacard, Martinet, Lachaux, Lacroix, Rozi{\`{e}}re, Goyal, Hambro, Azhar, Rodriguez, Joulin, Grave, and Lample}]{DBLP:journals/corr/abs-2302-13971}
Hugo Touvron, Thibaut Lavril, Gautier Izacard, Xavier Martinet, Marie{-}Anne Lachaux, Timoth{\'{e}}e Lacroix, Baptiste Rozi{\`{e}}re, Naman Goyal, Eric Hambro, Faisal Azhar, Aur{\'{e}}lien Rodriguez, Armand Joulin, Edouard Grave, and Guillaume Lample. 2023{\natexlab{a}}.
\newblock Llama: Open and efficient foundation language models.
\newblock \emph{CoRR}, abs/2302.13971.

\bibitem[{Touvron et~al.(2023{\natexlab{b}})Touvron, Martin, Stone, Albert, Almahairi, Babaei, Bashlykov, Batra, Bhargava, Bhosale, Bikel, Blecher, Canton{-}Ferrer, Chen, Cucurull, Esiobu, Fernandes, Fu, Fu, Fuller, Gao, Goswami, Goyal, Hartshorn, Hosseini, Hou, Inan, Kardas, Kerkez, Khabsa, Kloumann, Korenev, Koura, Lachaux, Lavril, Lee, Liskovich, Lu, Mao, Martinet, Mihaylov, Mishra, Molybog, Nie, Poulton, Reizenstein, Rungta, Saladi, Schelten, Silva, Smith, Subramanian, Tan, Tang, Taylor, Williams, Kuan, Xu, Yan, Zarov, Zhang, Fan, Kambadur, Narang, Rodriguez, Stojnic, Edunov, and Scialom}]{DBLP:journals/corr/abs-2307-09288}
Hugo Touvron, Louis Martin, Kevin Stone, Peter Albert, Amjad Almahairi, Yasmine Babaei, Nikolay Bashlykov, Soumya Batra, Prajjwal Bhargava, Shruti Bhosale, Dan Bikel, Lukas Blecher, Cristian Canton{-}Ferrer, Moya Chen, Guillem Cucurull, David Esiobu, Jude Fernandes, Jeremy Fu, Wenyin Fu, Brian Fuller, Cynthia Gao, Vedanuj Goswami, Naman Goyal, Anthony Hartshorn, Saghar Hosseini, Rui Hou, Hakan Inan, Marcin Kardas, Viktor Kerkez, Madian Khabsa, Isabel Kloumann, Artem Korenev, Punit~Singh Koura, Marie{-}Anne Lachaux, Thibaut Lavril, Jenya Lee, Diana Liskovich, Yinghai Lu, Yuning Mao, Xavier Martinet, Todor Mihaylov, Pushkar Mishra, Igor Molybog, Yixin Nie, Andrew Poulton, Jeremy Reizenstein, Rashi Rungta, Kalyan Saladi, Alan Schelten, Ruan Silva, Eric~Michael Smith, Ranjan Subramanian, Xiaoqing~Ellen Tan, Binh Tang, Ross Taylor, Adina Williams, Jian~Xiang Kuan, Puxin Xu, Zheng Yan, Iliyan Zarov, Yuchen Zhang, Angela Fan, Melanie Kambadur, Sharan Narang, Aur{\'{e}}lien Rodriguez, Robert Stojnic, Sergey Edunov,
  and Thomas Scialom. 2023{\natexlab{b}}.
\newblock Llama 2: Open foundation and fine-tuned chat models.
\newblock \emph{CoRR}, abs/2307.09288.

\bibitem[{Touvron et~al.(2023{\natexlab{c}})Touvron, Martin, Stone, Albert, Almahairi, Babaei, Bashlykov, Batra, Bhargava, Bhosale, Bikel, Blecher, Canton{-}Ferrer, Chen, Cucurull, Esiobu, Fernandes, Fu, Fu, Fuller, Gao, Goswami, Goyal, Hartshorn, Hosseini, Hou, Inan, Kardas, Kerkez, Khabsa, Kloumann, Korenev, Koura, Lachaux, Lavril, Lee, Liskovich, Lu, Mao, Martinet, Mihaylov, Mishra, Molybog, Nie, Poulton, Reizenstein, Rungta, Saladi, Schelten, Silva, Smith, Subramanian, Tan, Tang, Taylor, Williams, Kuan, Xu, Yan, Zarov, Zhang, Fan, Kambadur, Narang, Rodriguez, Stojnic, Edunov, and Scialom}]{llama2}
Hugo Touvron, Louis Martin, Kevin Stone, Peter Albert, Amjad Almahairi, Yasmine Babaei, Nikolay Bashlykov, Soumya Batra, Prajjwal Bhargava, Shruti Bhosale, Dan Bikel, Lukas Blecher, Cristian Canton{-}Ferrer, Moya Chen, Guillem Cucurull, David Esiobu, Jude Fernandes, Jeremy Fu, Wenyin Fu, Brian Fuller, Cynthia Gao, Vedanuj Goswami, Naman Goyal, Anthony Hartshorn, Saghar Hosseini, Rui Hou, Hakan Inan, Marcin Kardas, Viktor Kerkez, Madian Khabsa, Isabel Kloumann, Artem Korenev, Punit~Singh Koura, Marie{-}Anne Lachaux, Thibaut Lavril, Jenya Lee, Diana Liskovich, Yinghai Lu, Yuning Mao, Xavier Martinet, Todor Mihaylov, Pushkar Mishra, Igor Molybog, Yixin Nie, Andrew Poulton, Jeremy Reizenstein, Rashi Rungta, Kalyan Saladi, Alan Schelten, Ruan Silva, Eric~Michael Smith, Ranjan Subramanian, Xiaoqing~Ellen Tan, Binh Tang, Ross Taylor, Adina Williams, Jian~Xiang Kuan, Puxin Xu, Zheng Yan, Iliyan Zarov, Yuchen Zhang, Angela Fan, Melanie Kambadur, Sharan Narang, Aur{\'{e}}lien Rodriguez, Robert Stojnic, Sergey Edunov,
  and Thomas Scialom. 2023{\natexlab{c}}.
\newblock \href {https://doi.org/10.48550/ARXIV.2307.09288} {Llama 2: Open foundation and fine-tuned chat models}.
\newblock \emph{CoRR}, abs/2307.09288.

\bibitem[{Wang et~al.(2018)Wang, Singh, Michael, Hill, Levy, and Bowman}]{wang-etal-2018-glue}
Alex Wang, Amanpreet Singh, Julian Michael, Felix Hill, Omer Levy, and Samuel Bowman. 2018.
\newblock \href {https://doi.org/10.18653/v1/W18-5446} {{GLUE}: A multi-task benchmark and analysis platform for natural language understanding}.
\newblock In \emph{Proceedings of the 2018 {EMNLP} Workshop {B}lackbox{NLP}: Analyzing and Interpreting Neural Networks for {NLP}}, pages 353--355, Brussels, Belgium. Association for Computational Linguistics.

\bibitem[{Wang et~al.(2022)Wang, Kordi, Mishra, Liu, Smith, Khashabi, and Hajishirzi}]{self-instruct}
Yizhong Wang, Yeganeh Kordi, Swaroop Mishra, Alisa Liu, Noah~A. Smith, Daniel Khashabi, and Hannaneh Hajishirzi. 2022.
\newblock \href {https://doi.org/10.48550/arXiv.2212.10560} {Self-instruct: Aligning language model with self generated instructions}.
\newblock \emph{CoRR}, abs/2212.10560.

\bibitem[{Wei et~al.(2022{\natexlab{a}})Wei, Bosma, Zhao, Guu, Yu, Lester, Du, Dai, and Le}]{zeroshotlearner}
Jason Wei, Maarten Bosma, Vincent~Y. Zhao, Kelvin Guu, Adams~Wei Yu, Brian Lester, Nan Du, Andrew~M. Dai, and Quoc~V. Le. 2022{\natexlab{a}}.
\newblock \href {https://openreview.net/forum?id=gEZrGCozdqR} {Finetuned language models are zero-shot learners}.
\newblock In \emph{The Tenth International Conference on Learning Representations, {ICLR} 2022, Virtual Event, April 25-29, 2022}. OpenReview.net.

\bibitem[{Wei et~al.(2022{\natexlab{b}})Wei, Wang, Schuurmans, Bosma, Ichter, Xia, Chi, Le, and Zhou}]{cot}
Jason Wei, Xuezhi Wang, Dale Schuurmans, Maarten Bosma, Brian Ichter, Fei Xia, Ed~H. Chi, Quoc~V. Le, and Denny Zhou. 2022{\natexlab{b}}.
\newblock \href {http://papers.nips.cc/paper\_files/paper/2022/hash/9d5609613524ecf4f15af0f7b31abca4-Abstract-Conference.html} {Chain-of-thought prompting elicits reasoning in large language models}.
\newblock In \emph{NeurIPS}.

\bibitem[{Wu et~al.(2020)Wu, Kuang, Zhang, Liu, Sun, Xiao, Zhuang, Si, and Wu}]{wu2020biased}
Yiquan Wu, Kun Kuang, Yating Zhang, Xiaozhong Liu, Changlong Sun, Jun Xiao, Yueting Zhuang, Luo Si, and Fei Wu. 2020.
\newblock De-biased court’s view generation with causality.
\newblock In \emph{Proceedings of the 2020 Conference on Empirical Methods in Natural Language Processing (EMNLP)}, pages 763--780.

\bibitem[{Zeng et~al.(2023)Zeng, Liu, Du, Wang, Lai, Ding, Yang, Xu, Zheng, Xia, Tam, Ma, Xue, Zhai, Chen, Liu, Zhang, Dong, and Tang}]{DBLP:conf/iclr/ZengLDWL0YXZXTM23}
Aohan Zeng, Xiao Liu, Zhengxiao Du, Zihan Wang, Hanyu Lai, Ming Ding, Zhuoyi Yang, Yifan Xu, Wendi Zheng, Xiao Xia, Weng~Lam Tam, Zixuan Ma, Yufei Xue, Jidong Zhai, Wenguang Chen, Zhiyuan Liu, Peng Zhang, Yuxiao Dong, and Jie Tang. 2023.
\newblock {GLM-130B:} an open bilingual pre-trained model.
\newblock In \emph{ICLR}. OpenReview.net.

\bibitem[{Zeng et~al.(2020)Zeng, Li, Zhai, and Zhang}]{zeng2020counterfactual}
Xiangji Zeng, Yunliang Li, Yuchen Zhai, and Yin Zhang. 2020.
\newblock Counterfactual generator: A weakly-supervised method for named entity recognition.
\newblock In \emph{Proceedings of the 2020 Conference on Empirical Methods in Natural Language Processing (EMNLP)}, pages 7270--7280.

\bibitem[{Zhu et~al.(2020)Zhu, Zhang, Liu, and Wang}]{zhu2020counterfactual}
Qingfu Zhu, Weinan Zhang, Ting Liu, and William~Yang Wang. 2020.
\newblock Counterfactual off-policy training for neural dialogue generation.
\newblock In \emph{Proceedings of the 2020 Conference on Empirical Methods in Natural Language Processing (EMNLP)}, pages 3438--3448.

\end{thebibliography}
\bibliographystyle{acl_natbib}

\appendix



\end{document}